\documentclass{article}

\usepackage[final]{nips_2016}

\usepackage[utf8]{inputenc} % allow utf-8 input
\usepackage[T1]{fontenc}    % use 8-bit T1 fonts
\usepackage[draft]{hyperref}       % hyperlinks
\usepackage{url}            % simple URL typesetting
\usepackage{booktabs}       % professional-quality tables
\usepackage{amsfonts}       % blackboard math symbols
\usepackage{nicefrac}       % compact symbols for 1/2, etc.
\usepackage{microtype}      % microtypography

\usepackage{amsmath,graphicx}
\usepackage{bm}
\usepackage{url}
\usepackage{amsmath}
\usepackage{amsthm}
\usepackage{graphicx}
\usepackage{subcaption}
\usepackage{algorithm2e}
\usepackage{pbox}
\usepackage{multirow}

\title{Invariant Representations for Noisy Speech Recognition}
\author{
  Dmitriy Serdyuk\thanks{Dmitriy Serdyuk performed the work 
    during an internship at IBM Watson.} \\
  MILA,
  Universit\'e de Montr\'eal\\
  Montr\'eal, QC H3T 1J4 \\
  \texttt{serdyuk@iro.umontreal.ca} \\
  \And
  Kartik Audhkhasi \\
  IBM Watson, \\
  Yorktown Heights, NY \\
  \And
  Phil\'emon Brakel \\
  MILA\\
  Universit\'e de Montr\'eal\\
  \And
  Bhuvana Ramabhadran\\
  IBM Watson,
  Yorktown Heights, NY \\
  \And
  Samuel Thomas\\
  IBM Watson,
  Yorktown Heights, NY \\
  \And
  Yoshua Bengio\\
  MILA, CIFAR Fellow,
  Universit\'e de Montr\'eal
}

\begin{document}

\maketitle

\begin{abstract}
    Modern automatic speech recognition (ASR) systems need to be robust under acoustic variability arising from environmental, speaker, channel, and recording conditions. Ensuring such robustness to variability is a challenge in modern day neural network-based ASR systems, especially when all types of variability are not seen during training. We attempt to address this problem by encouraging the neural network acoustic model to learn invariant feature representations.
    We use ideas from recent research on image generation using
    Generative Adversarial Networks and domain adaptation ideas extending
    adversarial gradient-based training. A recent work from Ganin et al. proposes to
    use adversarial training for image domain adaptation by using an intermediate
    representation from the main target classification network to deteriorate the domain 
    classifier performance through a separate neural network.
    Our work focuses on investigating neural architectures which produce
    representations invariant to noise conditions for ASR.  We evaluate the proposed architecture on the Aurora-4 task, a popular benchmark for
    noise robust ASR. We show that our method generalizes better than the standard multi-condition training especially when only a few noise categories are seen during training.
\end{abstract}

\section{Introduction}
\label{sec:intro}
    One of the most challenging aspects of automatic speech recognition (ASR)
    is the mismatch between the training and testing acoustic conditions. During
    testing, a system may encounter new recording conditions, microphone types, speakers,
    accents and types of background noises. Furthermore, even if the test scenarios are seen during training, there can be significant variability in their statistics. Thus, its important to develop ASR systems that are invariant to unseen acoustic conditions.
    Several model and feature based adaptation methods such as Maximum Likelihood Linear Regression (MLLR),  feature-based MLLR 
    and iVectors~\citep{saon2013speaker} have been proposed to handle speaker variability; and Noise Adaptive Training \citep[NAT;][]{kalinli2010noise}
    and Vector Taylor Series \citep[VTS;][]{un1998speech} to handle environment variability. With the increasing success of Deep Neural Network (DNN) 
    acoustic models for ASR \citep{hinton2012deep,seide2011conversational,sainath2011making}, end-to-end systems are being proposed~\citep{miao2015eesen,sainath2015learning} 
    for modeling the acoustic conditions within a single network. This allows us to take advantage of the network's ability to learn 
    highly non-linear feature transformations, with greater flexibility in constructing training objective functions that promote learning 
    of noise invariant representations. The main idea of this work is to force the acoustic model to learn a representation invariant to 
    noise conditions, instead of explicitly using noise robust acoustic features (Section~\ref{sec:invariant-speech}). This type of 
    noise-invariant training requires noise-condition labels during training only. It is related to the idea of generative adversarial 
    networks (GAN) and the gradient reverse method proposed in~\cite{goodfellow2014generative} and~\cite{ganin2014unsupervised} respectively 
    (Section~\ref{sec:relatedwork}). We present results on the Aurora-4 speech recognition task in Section~\ref{sec:experiments} and summarize 
    our findings in Section~\ref{sec:discussion}.

\section{Related Work}
\label{sec:relatedwork}
\emph{Generative Adversarial Networks} consist of two networks: generator and discriminator. 
    The generator network $G$ has an
    input of randomly-generated feature vectors and is asked to produce a sample, e.g. an image, 
    similar to the images in the training set. The discriminator network $D$
    can either receive a generated image from the generator $G$ or an image
    from the training set. Its task is to distinguish
    between the ``fake'' generated image and the ``real'' image taken from the dataset. Thus,
    the discriminator is just a classifier network with a sigmoid output layer
    and can be trained with gradient backpropagation. This gradient can be propagated further
    to the generator network.

    Two networks in the GAN setup are competing with each other: the 
    generator is trying to deceive the discriminator network, while the discriminator tries
    to do its best to recognize if there was a deception, similar to adversarial game-theoretic settings.    
    Formally, the objective function of GAN training is
    \begin{align*}
        \min_G \max_D V(D, G) = \mathbb{E}_{\bm{x} \sim p_{\text{data}}(\bm{x})}[\log D(\bm{x})] + 
            \mathbb{E}_{\bm{z} \sim p_{\bm{z}}(\bm{z})}[\log (1 - D(G(\bm{z})))].
    \end{align*}
    The maximization over the discriminator $D$ forms a usual cross-entropy objective, the gradients are
    computed with respect to the parameters of $D$. The parameters of $G$ are minimized using the gradients
    propagated through the second term. The minimization over $G$ makes it to produce examples which $D$
    classifiers as the training ones.

    Several practical guidelines were proposed for optimizing GANs in~\cite{radford2015unsupervised} and 
    further explored in~\cite{salimans2016improved}.
    
    Prior work by~\cite{ganin2014unsupervised} proposed a method of training a network 
    which can be adapted to new domains. The training data consists of the images
    labeled with classes of interest and separate domain (image background) labels. 
    The network has a $Y$-like structure: the image is fed to the
    first network which produces a hidden representation $h$. Then this representation $h$ is input to two separate networks: a domain classifier network (D) and 
    a target classifier network (R). The goal of training is to learn the hidden 
    representation that is invariant to the domain labels and performs well on the target classification task, so that the domain information doesn't 
    interfere with the target classifier at test time. Similar to the GAN objective, which forces the generation distribution be close to the data distribution,
    the \emph{gradient reverse method} makes domain distributions similar to each other.

    The network is trained with three goals: the hidden representation $h$ should
    be helpful for the target classifier, harmful for the domain classifier,
    and the domain classifier should have a good classification accuracy. More 
    formally, the authors define the loss function as
    \begin{equation}
        L = L_1(\hat{y}, y; \theta_R, \theta_E) + 
        \alpha L_2(\hat{d}, d; \theta_D) -
        \beta L_3(\hat{d}, d; \theta_E),
        \label{eq:grm}
    \end{equation}
    where $y$ is the ground truth class, $d$ is the domain label, corresponding
    hat variables are the network predictions, and $\theta_E, \theta_R$ and $\theta_D$ are the subsets of  parameters for the encoder,
    recognizer and the domain classifier networks respectively. The hyper-parameters
    $\alpha$ and $\beta$ denote the relative influence of the loss functions terms.

The influence of representations produced by a neural network to internal noise reduction is discussed in~\cite{yu2013feature} and this work sets a baseline for experiments on  Aurora-4 dataset. Recently, in~\cite{yusuke2016adversarial} a multilayer sigmoidal network is trained in an adversarial fashion on an in-house transcription task corrupted by noise.
\section{Invariant Representations for Speech Recognition}
\label{sec:invariant-speech}

Most ASR systems are DNN-HMM hybrid systems. The context dependent (CD) HMM states (acoustic model) are the class labels of interest. The
recording conditions, speaker identity, or gender represent the domains in GANs. The task is to make the hidden layer representations of the HMM state classifier network 
invariant with respect to these domains. We hypothesize that this adversarial method of
training helps the HMM state classifier to generalize better to unseen domain conditions and requires only a  
small additional amount of supervision, i.e. the domain labels.  

\begin{figure}
    \centering
    \captionsetup[subfigure]{oneside,margin={0.3cm,0cm}}
    \begin{subfigure}[b]{0.3\linewidth}
        \centering
        \includegraphics[width=0.9\linewidth]{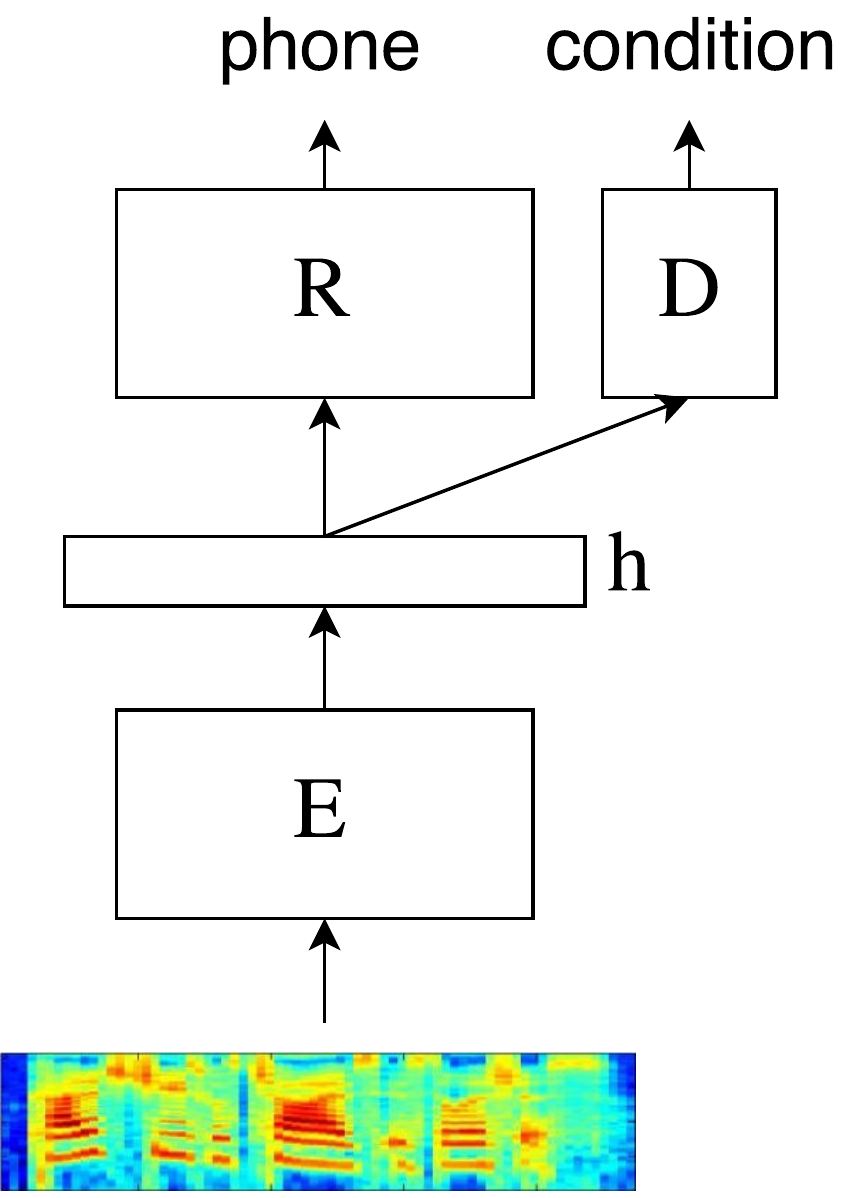}
        \caption{The model consists of three neural networks. The encoder $E$ produces
        the intermediate representation $h$ which used in the recognizer $R$ and 
        in the domain discriminator $D$. The hidden representation $h$ is trained to improve
        the recognition and minimize the domain discriminator accuracy. The domain discriminator
        is a classifier trained to maximize its accuracy on the noise type
        classification task.}
        \label{fig:model}
    \end{subfigure}%
    \begin{subfigure}[b]{0.7\linewidth}
        \centering
        \includegraphics[width=\linewidth]{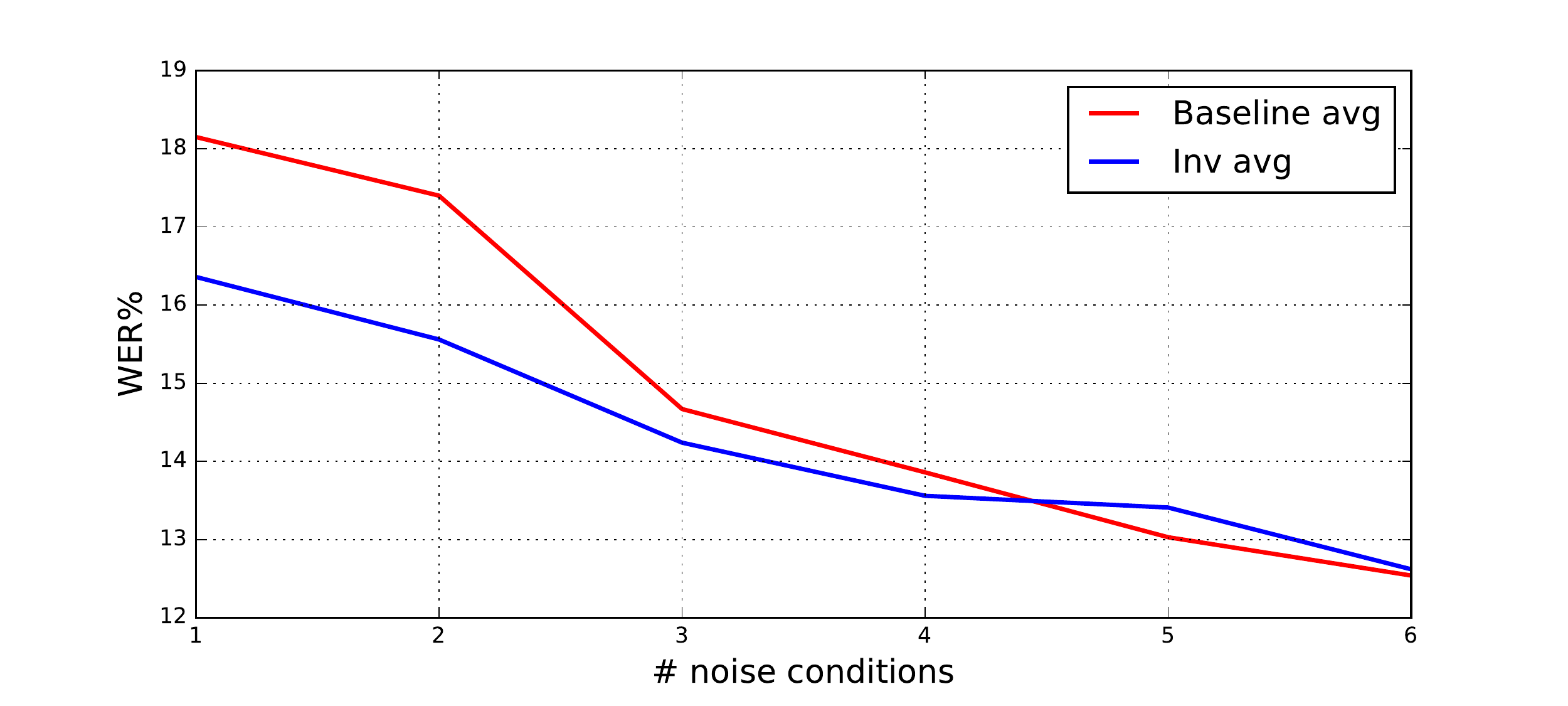}
        \includegraphics[width=\linewidth]{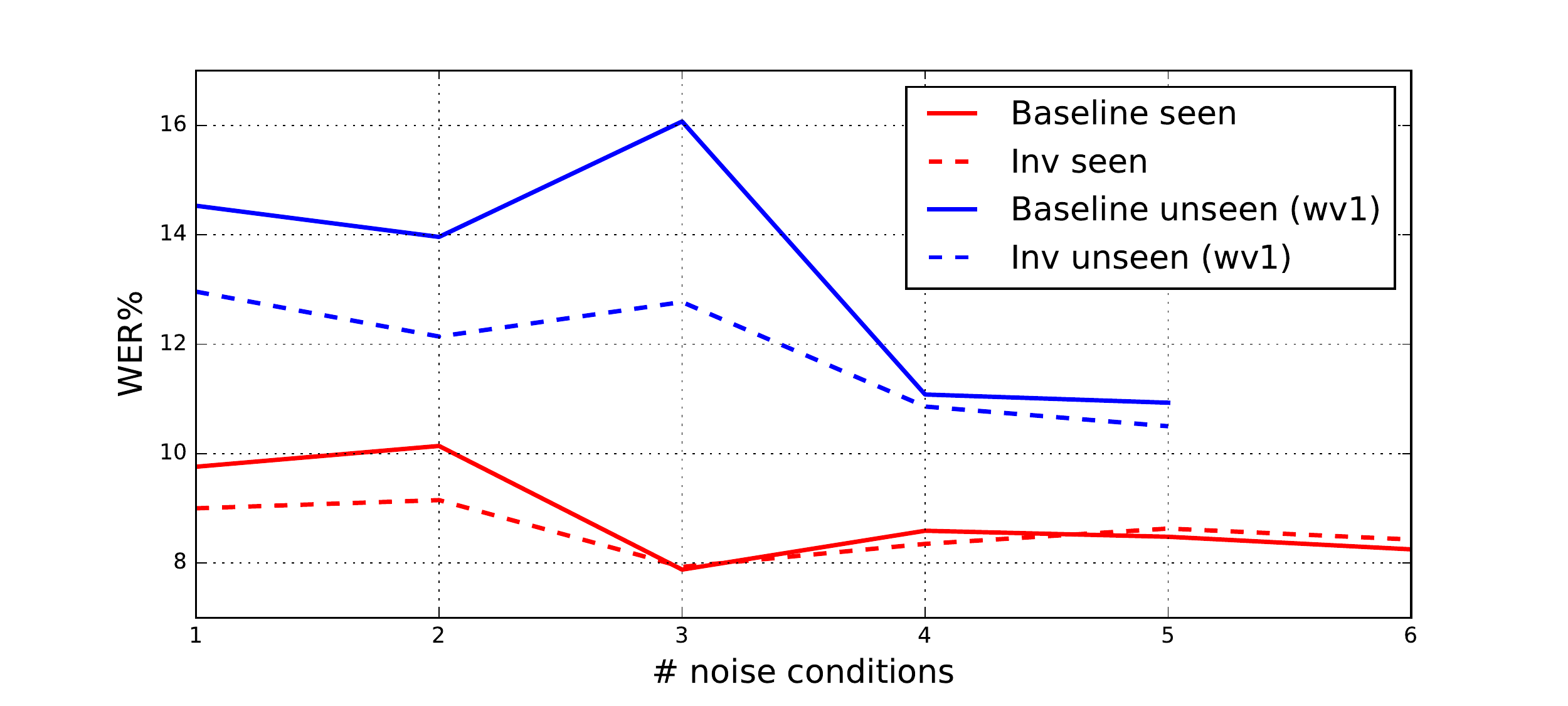}
        \caption{Up: Average performance of the baseline multi-condition and invariance model varying with  the number of noise
            conditions used for training. Bottom: Average performance on seen versus unseen noise conditions.
            Testing was performed on all wv1 conditions (Sennheiser microphone).
            }
        \label{fig:results}
    \end{subfigure}
    \caption{Model structure for invariant training and ASR results.}
\end{figure}

Figure~\ref{fig:model} depicts the model, which is same as the model for the gradient reverse method. It is a feed-forward neural network trained to predict the CD HMM state, with a branch that predicts the domain (noise condition). This branch is discarded in the testing phase. In our experiments we
used the noise condition as the domain label merging all noise types into one label
and clean as the other label. Our training loss function is Eq.~\ref{eq:grm} with 
$L_3$ set to $d\log(1 - \hat{d}) + (1-d)\log(\hat{d})$ for stability during training. 
$L_3$ term maximizes the probability
of an incorrect domain classification in contrast to the gradient reverse where the 
correct classification is minimized.
The terms $L_1$ and $L_2$ are 
regular cross-entropies which are minimized with corresponding parameters $\theta_E$ and $\theta_D$.
For simplicity, we use only a single hyper-parameter -- the weight of the third term.
 
\section{Experiments}
\label{sec:experiments}
We experimentally evaluate our approach   
on the well-benchmarked Aurora-4 \citep{parihar2002aurora} noisy speech recognition task. Aurora-4
is based on the Wall Street Journal corpus (WSJ0). It contains noises of 
six categories which was added to clean data. Every clean and noisy utterance is filtered to simulate the frequency characteristics. 
The training
data  contains 4400 clean utterances and 446 utterances for each noise condition,
i.e. a total of 2676 noisy utterances.
The test set consists of clean data, data corrupted by 6 noise types, and data recorded with a different microphone for both clean and noisy cases.

For both clean and noisy data, we extract 40-dimensional Mel-filterbank features with their deltas and 
delta-deltas spliced over $\pm$5 frames, resulting in 1320 input 
features that are subsequently mean and variance normalized.  The baseline acoustic model is a  6-layer 
DNN with 2048 rectified linear units at every layer. It is trained using momentum-accelerated stochastic gradient descent for 15 epochs with new-bob 
annealing~\citep[as in][]{morgan1995continuous,sainath2011making}.

In order to evaluate the impact of our method on generalization to unseen noises,
we performed 6 experiments with different set of seen noises. The networks are trained
on clean data, with each noise condition added one-by-one in the following order: airport, babble, car, 
restaurant, street, and train. The last training group includes all noises therefore matches the
standard multi-condition training setup. For every training group, we trained the
baseline and the invariance model where we branch out at the $4^{th}$ layer to an  
binary classifier predicting clean versus noisy data. Due to the imbalance between amounts of clean and
noisy utterances, we had to oversample noisy frames to ensure that every mini-batch contained
equal number of clean and noisy speech frames.

Table~\ref{tab:results} summarizes the results. Figure~\ref{fig:results} visualizes the word error rate for the baseline multi-condition training and invariance training as the number of seen noise types varies. We conclude that the best performance
gain is achieved when a small number of noise types are available during training. It can be seen that invariance training is able to generalize better to unseen noise types compared with multi-condition training.

We note that our experiments did not use layer-wise pre-training, commonly used for small
datasets. The baseline WERs reported are very close to the state-of-the-art. 
Our preliminary experiments on a pre-trained network (better overall WER) when using all noise types (last row of Table~\ref{tab:results}) for training show the same trend as the non-pretrained networks.

\begin{table}
    \centering
    \caption{Average word error rate (WER\%) on Aurora-4 dataset on all test conditions,
        including seen and unseen noise and unseen microphone. First column
        is the number of noise conditions used for the training. The last row is a 
        preliminary experiment with layer-wise pre-training close to state-of-the-art
        model and a corresponding invariance training starting with a pretrained model.}
    \label{tab:results}
    \begin{tabular}{r|cc||cc|cc|cc|cc}
        Noise       &Inv&BL&  \multicolumn{2}{c|}{A} & \multicolumn{2}{c|}{B} & \multicolumn{2}{c|}{C} & \multicolumn{2}{c}{D}\\
               & & &  Inv & BL & Inv & BL & Inv & BL & Inv & BL\\
    \hline
    1           &16.36        &18.14 &6.54&7.57    &12.71& 14.09   & 11.45&   13.10    & 22.47 &   24.80    \\
    2           &15.56        &17.39 &5.90&  6.58 &   11.69   &13.28   &11.12   &13.51   &21.79   &23.96 \\
    3           &14.24        &14.67 &5.45 & 5.08&    10.76&   12.44&   9.75&    9.84 &   19.93&   19.30\\
    4           &13.61        &13.84 & 5.08 &5.29    &9.73    &9.97    &9.49    &9.56    &19.49   &19.90\\         
    5           &13.41        &13.02 & 5.12 &5.34    &9.52    &9.42    &9.55    &8.67    &19.33   &18.65\\         
    6           &12.62        &12.60 & 4.80 &4.61    &9.04    &8.86    &8.76    &8.59    &18.16   &18.21\\
    \hline\hline
    6* &11.85        &11.99    &4.52    &4.76    &8.76    &8.76    &7.79    &8.57    &16.84&    16.99
    \end{tabular}
\end{table}

\section{Discussion}
\label{sec:discussion}
    This paper presents the application of generative adversarial networks and invariance training for noise robust speech recognition. We show that invariance training 
    helps the ASR system to generalize better to unseen noise conditions and improves word error rate when a small number of noise types are seen during training. Our 
    experiments show that in contrast to the image recognition task, in speech recognition,  the domain adaptation network suffers from underfitting. Therefore, the 
    gradient of the $L_3$ term in Eq.~\ref{eq:grm} is unreliable and noisy. Future research includes enhancements to the domain adaptation network while exploring alternative network architectures and invariance-promoting loss functions.
        
    %{\bf We need to make the message of this paragraph more clear.}
    % Dima: tried to fix -- it is differnt => needs more investigation

\section*{Acknowledgments}

We would like to thank Yaroslav Ganin, David Warde-Farley for insightful discussions,
developers of Theano~\cite{2016arXiv160502688short}, Blocks, and Fuel~\cite{MerrienboerBDSW15} 
for great toolkits.

%\bibliographystyle{IEEE}
\bibliographystyle{authordate1}
\bibliography{refs}

\end{document}